\documentclass[runningheads]{llncs}

\usepackage{enumitem}
\usepackage{amsmath}
\usepackage{algpseudocode}
\usepackage{algorithm}

\usepackage[T1]{fontenc}

\usepackage[labelfont=bf,font=small]{caption}
\usepackage[labelfont=bf,font=small]{subcaption}
\usepackage{graphicx}
\usepackage{hyperref}

\usepackage[separate-uncertainty=true]{siunitx}
\usepackage{booktabs,multirow}
\usepackage{xcolor}
\definecolor{graph1}{HTML}{B30033} 
\usepackage{tikz}
\usetikzlibrary{arrows,shapes,automata,backgrounds,petri,chains}
\usetikzlibrary{matrix,decorations.pathreplacing, calc, positioning,fit,trees}

\begin{document}
\title{Enhancing Bayesian Network Structural Learning with Monte Carlo Tree Search}
\titlerunning{Enhancing BN Structural Learning with MCTS}
%
\author{Jorge D. Laborda \inst{1,2}\orcidID{0000-0002-6844-3970} \and
Pablo Torrijos\inst{1,2}\orcidID{0000-0002-8395-3848} \and
Jos\'e M. Puerta\inst{1,2}\orcidID{0000-0002-9164-5191} \and
Jos\'e A. G\'amez\inst{1,2}\orcidID{0000-0003-1188-1117}}
\authorrunning{JD. Laborda, P. Torrijos, JM. Puerta and JA. G\'amez}
%
\institute{Instituto de Investigaci\'on en Inform\'atica de Albacete (I3A). Universidad de Castilla-La Mancha. Albacete, 02071, Spain. \and
Departamento de Sistemas Inform\'aticos. Universidad de Castilla-La Mancha. Albacete, 02071, Spain.\\
\email{\{JorgeDaniel.Laborda,Pablo.Torrijos,Jose.Puerta,Jose.Gamez\}@uclm.es}}
\maketitle              
\begin{abstract}
This article presents MCTS-BN, an adaptation of the Monte Carlo Tree Search (MCTS) algorithm for the structural learning of Bayesian Networks (BNs). Initially designed for game tree exploration, MCTS has been repurposed to address the challenge of learning BN structures by exploring the search space of potential ancestral orders in Bayesian Networks. Then, it employs Hill Climbing (HC) to derive a Bayesian Network structure from each order. In large BNs, where the search space for variable orders becomes vast, using completely random orders during the rollout phase is often unreliable and impractical. We adopt a semi-randomized approach to address this challenge by incorporating variable orders obtained from other heuristic search algorithms such as Greedy Equivalent Search (GES), PC, or HC itself. This hybrid strategy mitigates the computational burden and enhances the reliability of the rollout process. Experimental evaluations demonstrate the effectiveness of MCTS-BN in improving BNs generated by traditional structural learning algorithms, exhibiting robust performance even when base algorithm orders are suboptimal and surpassing the gold standard when provided with favorable orders.

\end{abstract}
%
%
%
%
%
\section{Introduction}

Bayesian Networks (BNs) \cite{Jensen_Nielsen,Koller_Friedman} are probabilistic graphical models that capture complex dependencies among variables, enabling the representation of uncertainty in specific domains. Known for their intuitive graphical structures, BNs facilitate symbolic analysis of variable relationships. In an era where interpretable models and causal modeling are gaining significance, BNs are considered cutting-edge technology. They find applications in various domains, from medical purposes \cite{McLachlan2020,Xie2022} to forest fire analysis \cite{Sevinc2020} or risk assessment \cite{fenton2018risk}, making them valuable for automated decision-making.

In addressing complex problems, the conventional approach of expertly crafting graphical structures and conditional probability tables for BNs becomes impractical \cite{Kjaerulff_Madsen}. The field of BN structural learning, which delves into discovering underlying probabilistic connections between variables in complex systems, has made notable strides over the years \cite{alonso_GES_2018,chickering_optimal_2002,gamez_learning_2011,FGES:2017}.

This work introduces \textit{MCTS-BN}, the Monte Carlo Tree Search (MCTS) \cite{Kocsis2006,wiechowski2022} application for BN structural learning. Initially designed for board games, MCTS shows promise due to its ability to balance exploration and exploitation. We adapt MCTS for BN structural learning by representing network variables in the search tree, looking for a suitable order for the variables to facilitate using a fast learning algorithm limited to a specific order, such as Hill Climbing (HC) \cite{gamez_learning_2011}. This is necessary due to the huge number of times that this algorithm has to be executed during the MCTS exploration.

However, BNs face the challenge of a vast search space for variable orders, $n!$ with $n$ variables. Random exploration during MCTS becomes inefficient and unreliable. To address this, we propose a hybrid strategy combining randomness with topological orders derived from standard greedy algorithms like HC itself, Greedy Equivalent Search (GES) \cite{chickering_optimal_2002}, or the PC algorithm \cite{spirtes93}. This approach enhances exploration efficiency and reliability.

Our findings underscore the effectiveness of this approach, particularly in scenarios where base algorithms yield suboptimal results and where MCTS has more room for improvement. We delve into how the results vary with the different base algorithms over the iterations. In addition, it is shown that the better the ancestral order with which it is initialized, the shorter the execution time.

This paper is organized as follows. Section \ref{sec:preliminaries} provides the necessary background, covering Bayesian Networks, their structural learning, and Monte Carlo Tree Search. In Section \ref{sec:MCTS-BNs}, our proposal for adapting MCTS for BN structural learning is explained in detail. Then, Section \ref{sec:experiments} details the methodology used for the experimental evaluation and analyses the obtained results. Finally, Section \ref{sec:conclusions} summarizes our contributions, and we discuss potential future research.

%
%
\section{Preliminaries} \label{sec:preliminaries}

\subsection{Bayesian Network} 

A Bayesian Network (BN), formally represented as $\mathcal{B}=(\mathcal{G},\mathcal{P})$, constitutes a probabilistic graphical model with two fundamental components: a Directed Acyclic Graph (DAG) denoted as $\mathcal{G}=(\mathcal{X},\mathcal{A})$, and a set of Conditional Probability Tables (CPTs) denoted as $\mathcal{P}$. 
The DAG $\mathcal{G}=(\mathcal{X},\mathcal{A})$ encapsulates the network structure, where $\mathcal{X}=\{X_1,\dots,X_n\}$ represents the variables of the problem domain, and $\mathcal{A} = \{X_i \rightarrow X_j \mid X_i, X_j \in \mathcal{X} \land X_i \neq X_j\}$ denotes the directed edges between them. The CPTs ($\mathcal{P}$) factorize the joint probability distribution $P(\mathcal{X})$ using graphical structure $\mathcal{G}$ and the Markov's condition: 
\begin{equation}
        P(\mathcal{X}) = P(X_1,\dots,X_n) = \prod_{i=1}^{n}P(X_i|pa_{\mathcal{G}}(X_i)),
\end{equation}
where $pa_{\mathcal{G}}(X_i)$ denotes the set of parents of $X_i$ in $\mathcal{G}$.

\subsection{Structural Learning of Bayesian Networks} 
Structural learning of Bayesian Networks (BNs) \cite{Scanagatta2019} is a crucial task involving discovering the optimal network topology to represent the underlying probabilistic relationships among variables accurately. Learning the structure of a BN is an NP-hard problem \cite{Scanagatta2019}. Therefore, heuristic methods are used when the dimensionality of the problem domain increases.

Two main groups of algorithms exist in the realm of structural learning for BNs. On one hand, constraint-based algorithms, like the PC algorithm \cite{spirtes93}, employ hypothesis testing to derive conditional independencies between variables based on the available data. On the other hand, score+search algorithms explore the space of potential BNs using a metric dependent on the data. In this context, we specifically consider discrete data.

Among the score+search algorithms, Hill Climbing (HC) \cite{gamez_learning_2011} and Greedy Equivalent Search (GES) \cite{chickering_optimal_2002} are notable. Both algorithms operate similarly, with the distinction that HC searches within the space of DAGs while GES explores the space of equivalence classes. This makes HC faster, but it generally yields inferior results. Additionally, Fast GES (fGES) \cite{FGES:2017} deserves mention as an enhancement of GES designed to better adapt to high-dimensional domains.

\subsection{Monte Carlo Tree Search} \label{subsec:MCTS} 
Monte Carlo Tree Search (MCTS) \cite{Kocsis2006,wiechowski2022} is a versatile best-first search algorithm designed for sequential decision-making tasks, combining the Monte Carlo simulation with tree search. Originally designed for game-playing scenarios, an area in which it has excelled, e.g., with AlphaGo \cite{Silver2016} and AlphaZero \cite{Silver2018}, MCTS has found applications in diverse domains such as molecular design \cite{Bryant2022,Kajita2020}, renewable energies \cite{Bai2022}, autonomous vehicles \cite{Mo2022} or optimization \cite{Labbe2020}. Its ability to efficiently navigate extensive search spaces while balancing exploration and exploitation positions it as a powerful tool in diverse domains.

In a formal context, MCTS is well-suited for problems modeled by a Markov Decision Process (MDP) \cite{howard1960}. A MDP is defined as a tuple $(\mathcal{S},\mathcal{A}_S,P_a,R_a)$ where:
\begin{itemize}
    \item $\mathcal{S} = \{S_0,\dots,S_n\}$ is the state space, i.e., a set of possible states. $S_0$ is distinguished as the initial state.
    \item $\mathcal{A}_S$ is the actions space, i.e., the set of possible actions available to perform in the state $S$.
    \item $P_a(S_i,S_j)$ is the transition function, i.e., the probability that action $a$ in state $S_i$ will lead to $S_j$. 
    \item $R_a(S)$ is the reward of reaching the state $S$ by the action $a$.
\end{itemize}

In MCTS, the state space $\mathcal{S}$ is explored by iteratively building a search tree. Each node in these trees represents a potential state or configuration $S$ of the problem, starting with the root node corresponding to the initial state $S_0$. In contrast, the edges between states correspond to each action $a_S$ successively done in each node $S$ to achieve the final state. The algorithm evaluates possible sequences of actions, assessing their outcomes through random sampling. 

During this exploration process, the search tree dynamically adapts based on the acquired information, including simulation rewards and the number of visits to each node. The method seeks to balance exploration and exploitation to select the best action in each moment, discovering promising branches of the search tree while refining evaluations of less explored ones. To achieve this balance, the algorithm normally employs the Upper Confidence bounds applied to Trees (UCT) \cite{Kocsis2006} formula:
\begin{equation} \label{eq:UCT}
UCT_a(S) = avg(R_a(S)) + C \cdot \sqrt{\frac{\ln N(S)}{N_a(S)}}
\end{equation}
where $UCT_a(S)$ is the score obtained when action $a$ is performed in state $S$; $avg(R_a(S))$ is the mean reward obtained thus far when performing $a$ in $S$; $C$ is a constant controlling the balance between exploration and exploitation (typically set to $\sqrt{2}$ by default); $N(S)$ is the total number of simulations in state $S$; and $N_a(S)$ is the number of simulations in $S$ when action $a$ is performed.

MCTS follows a four-step cycle of iterations that can be stopped at any time, returning the best solution found so far:
\begin{itemize}
    \item \textbf{Selection:} Starting from the root node, MCTS selects a path through the tree using a selection strategy, such as UCT, until it reaches a leaf node or a terminal state.
    \item \textbf{Expansion:} Upon reaching a leaf node, MCTS expands the tree by adding at least one child node, the result of applying a new action to the state.
    \item \textbf{Simulation (Rollout):} MCTS conducts a complete random simulation from the newly added node until reaching a terminal state to estimate the outcome of the problem.
    \item \textbf{Backpropagation:} The simulation results are backpropagated up the tree, from the leaf to the root node, updating the statistics of all middle nodes.
\end{itemize} 

Challenges in applying MCTS include fine-tuning parameters, handling large state spaces, and addressing computational demands. Researchers have developed various MCTS variants to suit the intrinsic characteristics of each problem, such as parallel MCTS \cite{Chaslot2008}, multi-objective MCTS \cite{Perez2015,Weng2021}, real-time MCTS \cite{Perez2015}, or using a graph instead of a tree as in Monte Carlo Graph Search (MCGS) \cite{leurent20a}.


%
%
\section{MCTS for Structural Learning of BNs} \label{sec:MCTS-BNs}
In this section, we provide an in-depth exploration of how Monte Carlo Tree Search is applied to the problem of structural learning in Bayesian Networks. 

\subsection{Searching topological orders} \label{subsec:topological} 
A topological order of a Bayesian network is an ordered list $\sigma = \langle X_i, \dots, X_j \rangle$ of the $n$ BN variables $\mathcal{X} = \{X_1, \dots, X_n\}$, such that all the parents of any $X \in \mathcal{X}$ precede $X$ in $\sigma$. Note that, in general, there are several topological orders for a given DAG. Within this modified version of MCTS, the algorithm's objective is to explore the topological order $\sigma$ of the BN variables that maximizes the score obtained by a structural learning algorithm constrained to follow $\sigma$. Since the problem of structural learning of BNs is NP-hard, commonly used algorithms are greedy methods that perform a local search in a subset of the state space. Therefore, providing an order $\sigma$ as similar as possible to the topological order induced by the underlying distribution of the data will allow us to obtain better solutions than those obtained by the unconstrained algorithm.

Our approach, MCTS-BN, employs a Hill Climbing algorithm constrained to a specific $\sigma$ \cite{li18aHC} due to its simplicity and fast execution time while maintaining good results. This choice is particularly relevant in our context as the algorithm needs to be executed numerous times (once in each rollout step of MCTS). As for the scoring metric, we utilize the Bayesian Dirichlet equivalent uniform (BDeu) score \cite{heckerman_learning_1995}, a widely adopted metric in the history of score-based BN structural learning algorithms \cite{alonso_GES_2018,chickering_optimal_2002,gamez_learning_2011}, owing to its properties that guarantee that is a locally consistent scoring criterion \cite{chickering_optimal_2002}.

In this case, the MCTS search space $\mathcal{S} = \{\sigma_0, \sigma_1, \dots\}$ will consist of each possible (partial) order of the variables. Note that while the search space size for complete orders is $n!$, the search space in the case of partial orders is much larger, $\sum_{k=0}^{n} k! \binom{n}{k}$. Let us denote $vars(\sigma)$ as the set of variables in $\sigma$. In our proposal, an action consists of adding a new variable to a state, i.e., a partial order $\sigma_i$. Thus, by abuse of the notation, we use sets of variables as sets of actions, being $A_{\sigma_i} = \mathcal{X} \setminus vars(\sigma_i)$ the actions set available at a state $\sigma_i$.

The method starts with the initial state $\sigma_0$ being an empty order $\langle \rangle$. New nodes (states) are added to the tree by applying a possible action $X_j \in A_{\sigma_i}$ over a node (state) $\sigma_i$, that is, by adding $X_i$ at the end of the partial order $\sigma_i$. For example, at the first level, the possible states (partial orders) will be $\{ \langle X_i \rangle \}_{i=1}^n$ since $\mathcal{A}_{\sigma_0} = \mathcal{X}$; at the second level, the state $\langle X_i \rangle$ is expanded to the set $\{ \langle X_i, X_j \rangle\}_{j\neq i}$; and so on for the rest of levels. When $\sigma$ contains the $n$ variables, it is a complete order and, therefore, a {\em final} state. 

During the rollout, an incomplete order $\sigma$ is evaluated by extending it with all the variables in $\mathcal{X} \setminus vars(\sigma)$. This creates a complete order $\sigma^+$. Subsequently, a BN is learned using an HC algorithm restricted to the topological ordering represented by $\sigma^+$, and the score attained by the obtained network is employed in the scoring process of $\sigma$.


\subsection{UCT adaptation} \label{subsec:uct} 
When applying MCTS for structural learning of BNs, adapting certain aspects is crucial to ensure that the UCT formula (Eq. \ref{eq:UCT}) yields useful values because of the different scales in the score that we use as reward $R_a(S)$ (BDeu) due to the dataset size and the network (domain). We manage these issues as follows:
\begin{itemize}
    \item Since the metric is proportional to the dataset size, we divide the BDeu score by the number of instances to deal with different dataset sizes. We denote this normalized BDeu value as $nBDeu$.
    
    \item As the scale of BDeu significantly differs between different domains, and it cannot be predicted from domain features (e.g. number of variables), we propose standardizing it once the algorithm has started. To do this, we have to test each possible action $\mathcal{A}_{\sigma_0} = \mathcal{X}$ at the initial state $\sigma_0$, so considering each variable as the first variable of order $\sigma$. This provides an overview of the nBDeu score scale for the current BN. Subsequently, the mean and standard deviation of the obtained scores are calculated and used to (zero-mean standardize) any reward computed during the algorithm's running, including those in this initial iteration. In this way, we ensure that a uniform scale is always maintained. We denote this standardized nBDeu as $snBDeu$.
\end{itemize}

Finally, adapting the UCT to the BN learning problem also involves adjusting the exploration constant $C$. The typical value of $\sqrt{2}$ for $C$ might be too large in this context, especially considering that the rewards have been standardized, resulting in smaller values. Using the standard $\sqrt{2}$ might overly prioritize exploration. We opt to set $C = \frac{\sqrt{2}}{100}$ as it strikes a good balance between exploration and exploitation. Therefore, the modified UCT algorithm will be:
\begin{equation}
UCT_a(S) = avg(snBDeu_a(S)) + \frac{\sqrt{2}}{100} \cdot \sqrt{\frac{\ln N(S)}{N_a(S)}}
\end{equation}

\subsection{Guided search} \label{subsec:guided} 

As aforementioned, a significant challenge in adapting MCTS for structural learning of BNs is the vast search space $\mathcal{S}$, where most of the $n!$ final states may lead to far from optimal solutions, making it difficult for MCTS to discover good orderings (and BNs) efficiently. Incorporating algorithms like HC in the MCTS evaluations becomes intricate, making it impossible to feasibly run millions or billions of iterations as seen in other MCTS implementations \cite{Silver2016,Silver2018}. Therefore, starting from a knowledge base to exploit existing information becomes particularly important to make the most of the iterations carried out. 

\begin{figure}[tb]
    \centering
    \begin{subfigure}{.3\textwidth}
        \centering
        \begin{tikzpicture}[->,>=stealth',shorten >=1pt,auto,semithick, node distance=1.1cm]
          \tikzstyle{every state}=[fill=graph1,draw=none,text=white]
                \node[state,inner sep=2pt,minimum size=3pt, fill=white, draw=graph1, text=graph1]  (A)  {$X_1$};
                \node[state,inner sep=2pt,minimum size=3pt]  (B) [below left of= A, fill=white, draw=graph1, text=graph1] {$X_4$};
                \node[state,inner sep=2pt,minimum size=3pt]  (C) [below right of= A, fill=white, draw=graph1, text=graph1] {$X_7$};
                \node[state,inner sep=2pt,minimum size=3pt]  (D) [below left of= C, fill=white, draw=graph1, text=graph1] {$X_2$};
                \node[state,inner sep=2pt,minimum size=3pt]  (E) [below right of= C, fill=white, draw=graph1, text=graph1] {$X_5$};
                \node[state,inner sep=2pt,minimum size=3pt]  (F) [below left of= D, fill=white, draw=graph1, text=graph1] {$X_3$};
                \node[state,inner sep=2pt,minimum size=3pt]  (G) [below right of= D, fill=white, draw=graph1, text=graph1] {$X_6$};
    
                \path (A) edge              node {} (B)
                      (A) edge              node {} (C)
                      (C) edge              node {} (D)
                      (D) edge              node {} (E)
                      (E) edge              node {} (G)
                      (D) edge              node {} (F)
                      (D) edge              node {} (G);
        \end{tikzpicture}
        \caption{BN learned with GES.}
        \label{subfig:BN}
    \end{subfigure}%
    \begin{subfigure}{.7\textwidth}
    \centering
        \begin{tikzpicture}[->,>=stealth',shorten >=1pt,auto,semithick, node distance=1.1cm]
          \tikzstyle{every state}=[fill=graph1,draw=none,text=white]
                \node[state,inner sep=2pt,minimum size=3pt, fill=white, draw=graph1, text=graph1, label=180:{$\sigma_{GES}$:\ \ \ }]  (A)  {$X_1$};
                \node[state,inner sep=2pt,minimum size=3pt]  (B) [right of= A, fill=white, draw=graph1, text=graph1] {$X_4$};
                \node[state,inner sep=2pt,minimum size=3pt]  (C) [right of= B, fill=white, draw=graph1, text=graph1] {$X_7$};
                \node[state,inner sep=2pt,minimum size=3pt]  (D) [right of= C, fill=white, draw=graph1, text=graph1] {$X_2$};
                \node[state,inner sep=2pt,minimum size=3pt]  (E) [right of= D, fill=white, draw=graph1, text=graph1] {$X_5$};
                \node[state,inner sep=2pt,minimum size=3pt]  (F) [right of= E, fill=white, draw=graph1, text=graph1] {$X_3$};
                \node[state,inner sep=2pt,minimum size=3pt]  (G) [right of= F, fill=white, draw=graph1, text=graph1] {$X_6$};
    
                \path (A) edge              node {} (B)
                      (B) edge              node {} (C)
                      (C) edge              node {} (D)
                      (D) edge              node {} (E)
                      (E) edge              node {} (F)
                      (F) edge              node {} (G);
        \end{tikzpicture}
    
        \vspace{0.2cm}%
    
        \begin{tikzpicture}[->,>=stealth',shorten >=1pt,auto,semithick, node distance=1.1cm]
          \tikzstyle{every state}=[fill=graph1,draw=none,text=white]
                \node[state,inner sep=2pt,minimum size=3pt, label=180:{\ \ \ \ \ $\sigma$:\ \ \ }]  (A)  {$X_5$};
                \node[state,inner sep=2pt,minimum size=3pt]  (B) [right of= A] {$X_1$};
                \node[state,inner sep=2pt,minimum size=3pt]  (C) [right of= B] {$X_2$};
                \node[state,inner sep=2pt,minimum size=3pt]  (D) [right of= C, fill=white, draw=white, text=white] {$X_1$};
                \node[state,inner sep=2pt,minimum size=3pt]  (E) [right of= D, fill=white, draw=white, text=white] {$X_1$};
                \node[state,inner sep=2pt,minimum size=3pt]  (F) [right of= E, fill=white, draw=white, text=white] {$X_1$};
                \node[state,inner sep=2pt,minimum size=3pt]  (G) [right of= F, fill=white, draw=white, text=white] {$X_1$};
    
                \path (A) edge              node {} (B)
                      (B) edge              node {} (C)
                      (C) edge [draw=white]              node {} (D)
                      (D) edge [draw=white]              node {} (E)
                      (E) edge [draw=white]              node {} (F)
                      (F) edge [draw=white]              node {} (G);
        \end{tikzpicture}
    
        \vspace{0.2cm}%
    
        \begin{tikzpicture}[->,>=stealth',shorten >=1pt,auto,semithick, node distance=1.1cm]
          \tikzstyle{every state}=[fill=graph1,draw=none,text=white]
                \node[state,inner sep=2pt,minimum size=3pt, label=180:{\ \ \ $\sigma^+$:\ \ \ }]  (A)  {$X_5$};
                \node[state,inner sep=2pt,minimum size=3pt]  (B) [right of= A] {$X_1$};
                \node[state,inner sep=2pt,minimum size=3pt]  (C) [right of= B] {$X_2$};
                \node[state,inner sep=2pt,minimum size=3pt]  (D) [right of= C, fill=white, draw=graph1, text=graph1] {$X_4$};
                \node[state,inner sep=2pt,minimum size=3pt]  (E) [right of= D, fill=white, draw=graph1, text=graph1] {$X_7$};
                \node[state,inner sep=2pt,minimum size=3pt]  (F) [right of= E, fill=white, draw=graph1, text=graph1] {$X_3$};
                \node[state,inner sep=2pt,minimum size=3pt]  (G) [right of= F, fill=white, draw=graph1, text=graph1] {$X_6$};
    
                \path (A) edge              node {} (B)
                      (B) edge              node {} (C)
                      (C) edge              node {} (D)
                      (D) edge              node {} (E)
                      (E) edge              node {} (F)
                      (F) edge              node {} (G);
        \end{tikzpicture}
        \caption{Topological orders.}
        \label{subfig:sigmas}
    \end{subfigure}

    \caption{Example of $\sigma^+$ construction from $\sigma$ and a $\sigma_{GES}$ from a BN.}
    \label{fig:sigmas}
    
\end{figure}
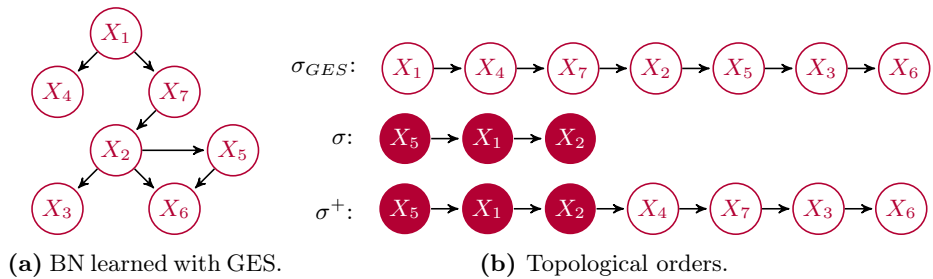

For this purpose, before running the MCTS algorithm, an unconstrained greedy structural learning algorithm, such as, e.g., GES, PC, or HC, is executed to discover an initial BN. Then, several topological orders are sampled from the learned BN to achieve diversity. We denote these orders as $\sigma_{GES}, \sigma_{HC}$ or $\sigma_{PC}$, depending on the learning algorithm employed. For example, let us assume that the BN shown in Figure \ref{subfig:BN} has been learned using the GES algorithm. Then, different $\sigma_{GES}$ orders extracted from it are $\langle X_1,X_7,X_2,X_3,X_5,X_6,X_4 \rangle$, $\langle X_1,X_7,X_4,X_2,X_5,X_3,X_6\rangle$, $\langle X_1,X_4,X_7,X_2,X_5,X_3,X_6 \rangle$, etc.

These orders serve two primary purposes. Firstly, they guide the process of completing incomplete orders for the rollout, that is, obtaining $\sigma^+$ from $\sigma$ in an informed way instead of randomly. Thus, given $\sigma$ to be complete and e.g. $\sigma_{GES}$ sampled from the learnt BN, we obtain $\sigma^+ = \sigma \cdot \sigma_{GES}^{\downarrow {\cal{X}}\setminus vars(\sigma)}$, where $\cdot$ is the concatenation operator and $\sigma_i^{\downarrow A}$ produces the projection of $\sigma_i$ over $A$, that is, the partial order obtained by removing from $\sigma$ the variables not in $A$. Following the example in Figure \ref{subfig:sigmas}, if $\sigma = \langle X_5, X_1, X_2 \rangle$, and $\sigma_{GES} = \langle X_1,X_4,X_7,X_2,X_5,X_3,X_6 \rangle$, then $\sigma^+ = \langle X_5, X_1, X_2 \rangle \cdot \langle X_1,X_4,X_7,X_2,X_5,X_3,X_6 \rangle^{\downarrow \{X_3, X_4, X_6, X_7\}}$ = $\langle X_5, X_1, X_2 \rangle \cdot \langle X_4,X_7,X_3,X_6\rangle$ = $\langle X_5, X_1, X_2, X_4,X_7,X_3,X_6\rangle$. This avoids closing promising paths due to bad luck by generating a bad random order.

The second use of these BN-sampled topological orders is for node expansion. Whether we consistently follow the same order of possible actions or opt for a random one each time we expand a state $\sigma$, there's a risk of selecting sink nodes very early in the order, leading to significantly suboptimal scores. This, in turn, reduces the likelihood of revisiting the expansion of state $\sigma$ even though it might be very promising. To avoid this, each state $\sigma$ will have attached a $\sigma_{GES}$ order that dictates the sequence in which each of its $\mathcal{A}_{\sigma}$ actions will be expanded.

%
%
\section{Experimental evaluation} \label{sec:experiments} 

\subsection{Methodology} \label{subsec:methodology}  
The following outlines our methodology for evaluating the MCTS-BN algorithm. We selected 6 real-world BNs from {\sf bnlearn}'s Bayesian Network Repository\footnote{\href{https://www.bnlearn.com/bnrepository/}{https://www.bnlearn.com/bnrepository/}}\cite{bnlearn}, three of small size  (\textit{Alarm}, \textit{Barley}, and \textit{Hepar2}) and three of very large size (\textit{Diabetes}, \textit{Link}, and \textit{Munin}). Each original network is considered as the {\em gold standard} for the corresponding domain. Subsequently, we sampled each BN to generate 10 datasets for each network, each one having 5000 instances. 

\begin{table}[htb]
\caption{Bayesian networks used in the experiments.}
\label{tabla-BNs}
\resizebox{\textwidth}{!} {%
\begin{tabular*}{1.3\textwidth}{@{\extracolsep{\fill}}lS[table-format=4.0]S[table-format=4.0]S[table-format=7.0]S[table-format=1.0]S[table-format=1.2]}
\toprule
\multicolumn{1}{c}{\multirow{2}{*}{\textsc{\bfseries Network}}} &\multicolumn{5}{c}{\textsc{\bfseries Features}} \\
\cmidrule(){2-6}
& \multicolumn{1}{r}{\textsc{\# Nodes}} & \multicolumn{1}{r}{\textsc{\# Edges}} & \multicolumn{1}{r}{\textsc{\# Parameters}} & \multicolumn{1}{r}{\textsc{Max. parents}} & \multicolumn{1}{r}{\textsc{Avg. degree}}\\
\midrule
\textsc{Alarm}           & 37	& 46 & 509 & 4 & 2.49\\
\textsc{Barley}          & 48	& 84 & 114005 & 4 & 3.5\\
\textsc{Hepar2}          & 70	& 123 & 1453 & 6 & 3.51\\
\midrule
\textsc{Diabetes}        & 413 & 602 & 429409 & 2 & 2.92\\
\textsc{Link}            & 724 & 1125 & 14211 & 3 & 3.11\\
\textsc{Munin}           & 1041 & 1397 & 80592 & 3 & 2.68\\
\bottomrule
\end{tabular*}
}
\end{table}

The evaluation metrics for assessing the quality of solutions include the BDeu score and the execution time averaged over the runs for the 10 datasets sampled for each BN. These metrics are assessed for both the structural learning algorithms (GES, fGES, PC, and HC), which establish the foundation for obtaining the orders utilized by MCTS, and are also measured at each of the 10000 iterations carried out by the MCTS algorithm.  It is important to note that these 10000 iterations are counted after the first complete expansion, which is always carried out from the initial state $\sigma_0$, and that is depicted separately in the graphs shown in Section \ref{subsec:results}. Furthermore, a topological order sampled from each gold standard BN is used to run an HC algorithm restricted to it. The outcome of this execution should be the optimal network as it uses the (an) actual topological ordering among the variables, thus it constitute an excellent model to compare with. As shown in our experiments, our algorithm sometimes obtains a better value because when considering a limited data set, the gold standard may no longer be the optimal model for that data.

\subsection{Reproducibility} \label{subsec:reproducibility}  
All code has been implemented using Java (OpenJDK 17) and the causal reasoning library Tetrad 7.1.2-2\footnote{\href{https://github.com/cmu-phil/tetrad/releases/tag/v7.1.2-2}{https://github.com/cmu-phil/tetrad/releases/tag/v7.1.2-2}}. The fGES and PC implementations used are those of Tetrad. In contrast, the GES implementation corresponds to an improved version \cite{alonso_GES_2018}. HC and MCTS have been implemented from scratch. To ensure the reproducibility of the experiments, all of the datasets and the code are provided at GitHub \footnote{\href{https://github.com/ptorrijos99/MCTS-BN}{https://github.com/ptorrijos99/MCTS-BN}}. In addition, all the datasets generated for each BN are available at OpenML\footnote{\href{https://www.openml.org/search?type=data\&uploader\_id=\%3D\_33148\&tags=bnlearn}{https://www.openml.org/search?type=data\&uploader\_id=\%3D\_33148\&tags=bnlearn}}. Regarding hardware, all the experiments were performed on machines with Intel Xeon E5-2650 8-Core Processors with 64 GB of RAM per execution.

\subsection{Results} \label{subsec:results}  

\begin{figure*}[tb]
    \centerline{\includegraphics[trim={0.6cm 0.5cm 0.35cm 0},clip,width=\linewidth]{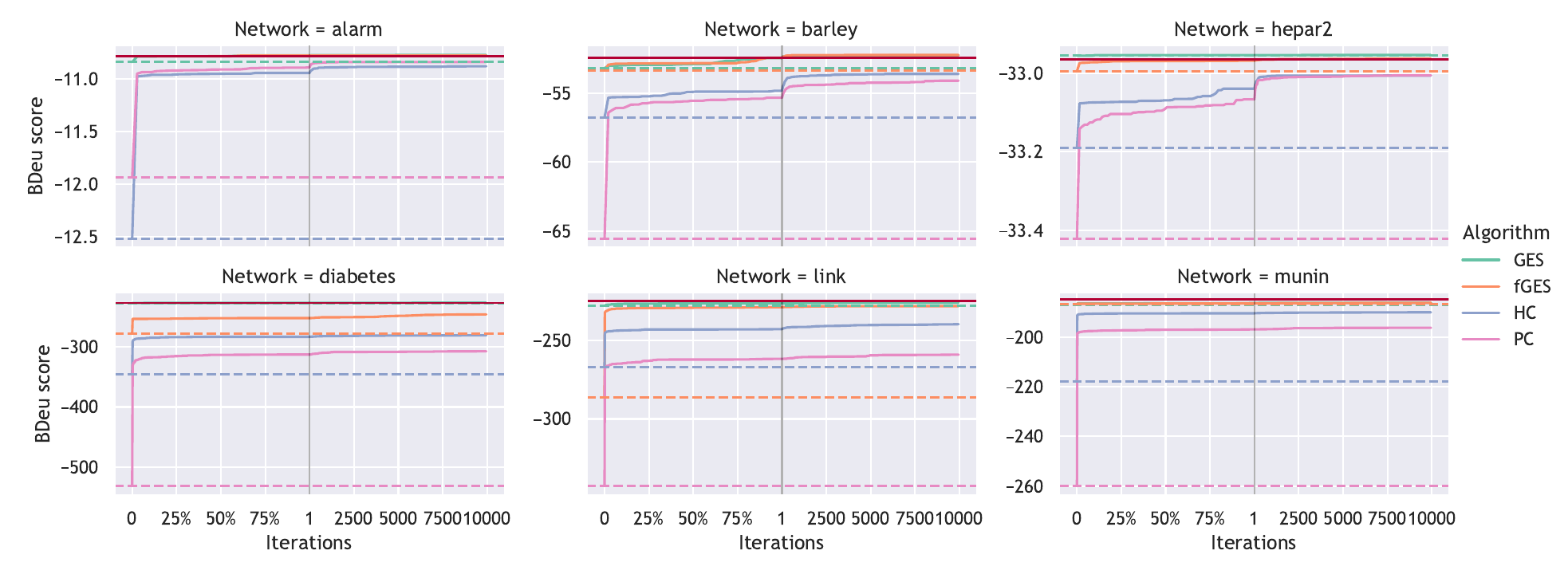}}
    \caption{\textit{nBDeu} score obtained by MCTS-BN over the iterations.}
    \label{fig:evolucion}
\end{figure*}

Figure \ref{fig:evolucion} illustrates the evolution of the BDeu score for MCTS-BN across iterations using the GES, fGES, PC, and HC algorithms as a basis. The left half of the figure showcases the evolution during the complete expansion of $\sigma_0$ as a percentage relative to the number of variables in each BN. In contrast, the right half displays the BDeu evolution over the 10000 MCTS iterations performed. Additionally, a dashed line represents the BDeu obtained by each algorithm used as a basis, accompanied by a red solid line marking the HC result restricted to the order obtained from the gold standard. 

\begin{figure*}[tb]
    \centerline{\includegraphics[trim={0.6cm 0.5cm 0.35cm 0},clip,width=\linewidth]{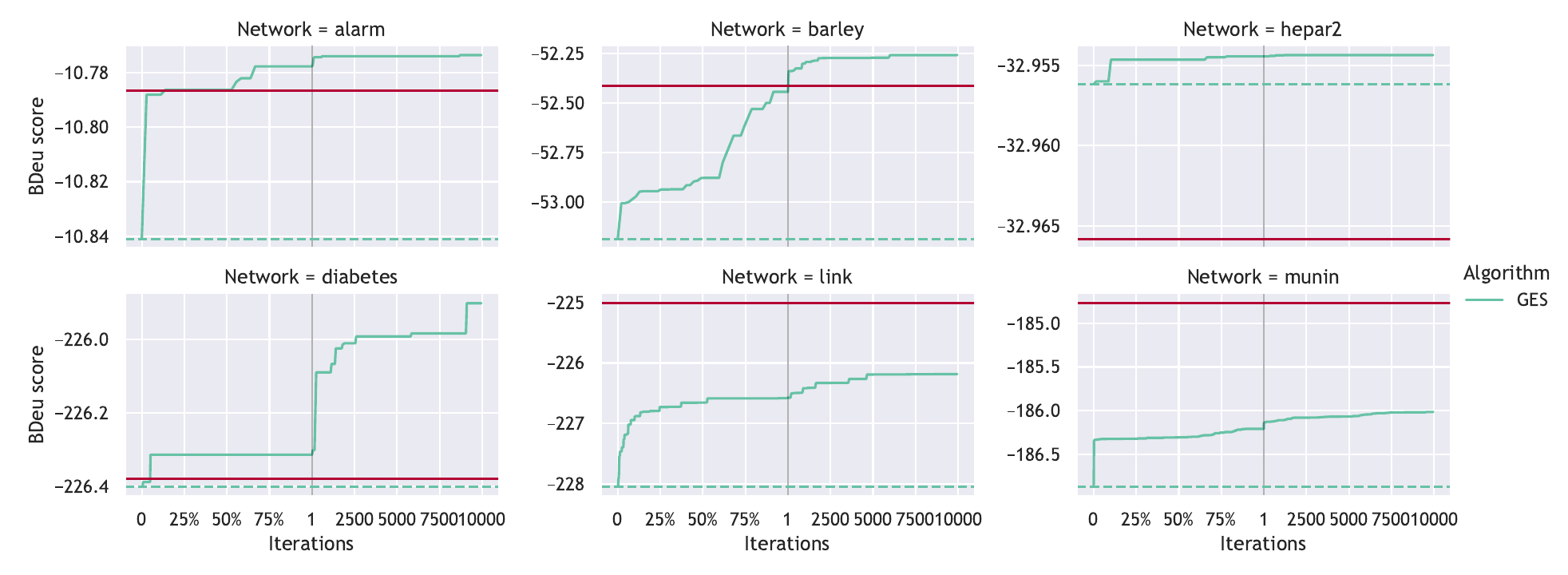}}
    \caption{\textit{nBDeu} score obtained by MCTS-BN over the iterations (GES).}
    \label{fig:evolucionGES}
\end{figure*}

Given the considerable variability in the scale of the results for the different algorithms, we present Figures \ref{fig:evolucionGESfGES} and \ref{fig:evolucionGES}, which depict the same data but exclusively for the GES and fGES algorithms and only for GES, respectively. This presentation can be understood as a kind of zoom and allows for a clearer appreciation of the evolution using these specific algorithms as a basis. The results lead to the following conclusions:

\begin{figure*}[tb]
    \centerline{\includegraphics[trim={0.6cm 0.5cm 0.35cm 0},clip,width=\linewidth]{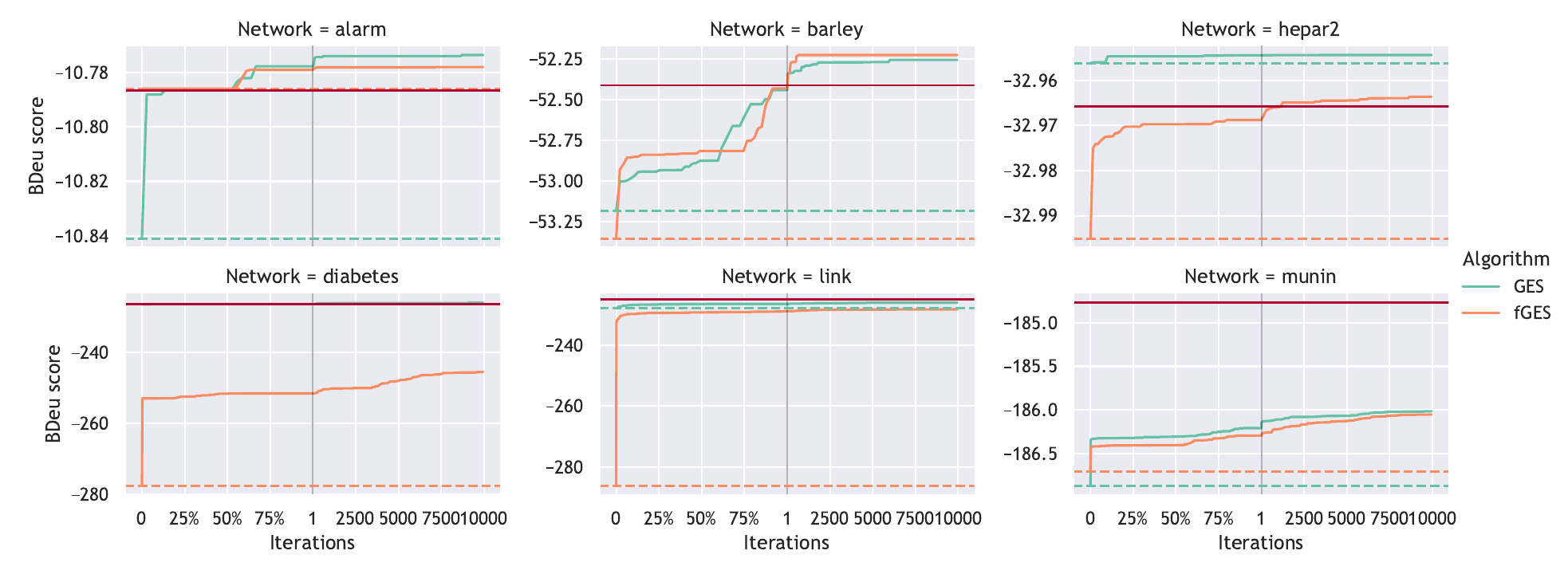}}
    \caption{\textit{nBDeu} score obtained by MCTS-BN over the iterations (GES, fGES).}
    \label{fig:evolucionGESfGES}
\end{figure*}

\begin{itemize}
    \item The use of MCTS-BN consistently proves beneficial across all base networks and algorithms used, notably enhancing the starting BDeu score results.

    \item In cases where the base algorithm generates suboptimal networks (as observed with HC, PC, and fGES in \textit{Diabetes} and \textit{Link}), MCTS-BN exhibits significant improvement potential. However, achieving results comparable to a good basic structural learning algorithm like GES becomes more challenging with worse starting conditions, particularly in larger BNs.

    \item In scenarios with small performance differences, such as in \textit{Alarm} with fGES outperforming GES, MCTS-BN demonstrates the capability to achieve similar final results regardless of the chosen base algorithm.

    \item MCTS-BN, when coupled with a robust base algorithm like GES, performs exceptionally well, surpassing the gold standard's upper bound in 4/6 BNs.
\end{itemize}

\begin{figure*}[tb]
    \centerline{\includegraphics[trim={0.4cm 0.5cm 0.35cm 0},clip,width=\linewidth]{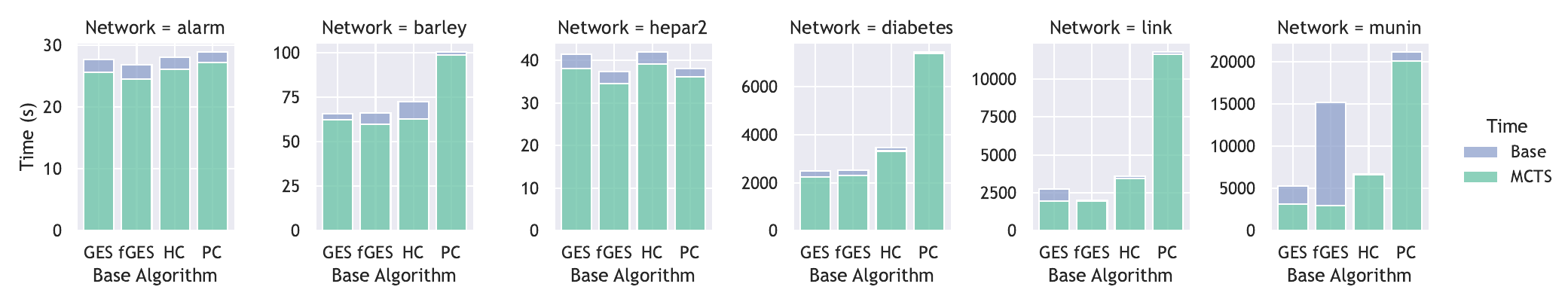}}
    \caption{Execution time (seconds) of the base algorithms and the posterior MCTS-BN.}
    \label{fig:time}
\end{figure*}

Obviously, the goal of this method is not to compare CPU time with greedy algorithms. However, we find it of interest to analyze it. Figure \ref{fig:time} shows the average times for each BN using GES, fGES, HC, and PC as the base algorithm. Each bar is further divided between the time of the base algorithm itself and that of the MCTS search. It illustrates that the MCTS time remains reasonable even with over 10000 runs of the limited HC. Additionally, it highlights the impact of the order generated by the base structural learning algorithm, especially evident in \textit{Diabetes}, \textit{Link}, and \textit{Munin} with HC and PC as the base algorithms. Thus, GES emerges not only as the algorithm with the best results but also as the one with the most favorable ratio of these results concerning execution time.

%
%
\section{Conclusions} \label{sec:conclusions}
Our study presents MCTS-BN, an adaptation of the Monte Carlo Tree Search for structural learning in Bayesian Networks. MCTS-BN conducts exploration within the search space of potential ancestral orders of the BN. A fast Hill Climbing constrained to a given order is executed in each iteration to generate a BN. We introduced an efficient exploration framework by utilizing predefined orders from other BN structural learning algorithms, such as HC, GES, fGES, or PC. This framework guides the algorithm towards favorable solutions, proving crucial in high-dimensional BNs where performing millions of iterations is challenging.

Our findings demonstrate that MCTS-BN generates high-quality BNs, achieving commendable results even when the orders from the underlying algorithms are suboptimal. MCTS-BN excels, surpassing the gold standard's results when provided with good orders. Looking forward, we envision exploring alternative base algorithms, integrating controlled randomness in node selection, and delving into distributed and federated versions of MCTS-BN.


\subsubsection{Acknowledgements} The following projects have funded this work: SBPLY/21/180225/000062 by the Government of Castilla-La Mancha and ``ERDF A way of making Europe''; PID2022-139293NB-C32, TED2021-131291B-I00 and FPU21/01074 by MCIN/AEI/10.13039/501100011033 and ``ESF Investing your future''; 2022-GRIN-34437 and 2019-PREDUCLM-10188 by Universidad de Castilla-La Mancha and ERDF funds.

This preprint has not undergone peer review or any post-submission improvements or corrections. The Version of Record of this contribution is published in Lecture Notes in Networks and Systems, vol 1174, and is available online at \href{https://doi.org/10.1007/978-3-031-74003-9\_32}{https://doi.org/10.1007/978-3-031-74003-9\_32}.

%
%
%
\bibliographystyle{splncs04}
\bibliography{biblio}

\begin{thebibliography}{10}
\providecommand{\url}[1]{\texttt{#1}}
\providecommand{\urlprefix}{URL }
\providecommand{\doi}[1]{https://doi.org/#1}

\bibitem{alonso_GES_2018}
Alonso, J.I., de~la Ossa, L., Gámez, J.A., Puerta, J.M.: On the use of local search heuristics to improve {GES}-based {Bayesian} network learning. Applied Soft Computing  \textbf{64},  366--376 (Mar 2018)

\bibitem{Bai2022}
Bai, F., Ju, X., Wang, S., Zhou, W., Liu, F.: Wind farm layout optimization using adaptive evolutionary algorithm with Monte Carlo Tree Search reinforcement learning. Energy Conversion and Management  \textbf{252},  115047 (Jan 2022)

\bibitem{Bryant2022}
Bryant, P., Pozzati, G., Zhu, W., Shenoy, A., Kundrotas, P., Elofsson, A.: Predicting the structure of large protein complexes using AlphaFold and Monte Carlo tree search. Nature Communications  \textbf{13}(1) (Oct 2022)

\bibitem{Chaslot2008}
Chaslot, G.M.J.B., Winands, M.H.M., van~den Herik, H.J.: Parallel Monte-Carlo Tree Search, p. 60–71. Springer Berlin Heidelberg (2008)

\bibitem{chickering_optimal_2002}
Chickering, D.M.: Optimal {Structure} {Identification} {With} {Greedy} {Search}. Journal of Machine Learning Research  \textbf{3}(Nov),  507--554 (2002)

\bibitem{fenton2018risk}
Fenton, N., Neil, M.: Risk Assessment and Decision Analysis with Bayesian Networks. Chapman and Hall/CRC (Sep 2018)

\bibitem{gamez_learning_2011}
Gámez, J.A., Mateo, J.L., Puerta, J.M.: Learning {Bayesian} networks by hill climbing: efficient methods based on progressive restriction of the neighborhood. Data Mining and Knowledge Discovery  \textbf{22}(1),  106--148 (Jan 2011)

\bibitem{heckerman_learning_1995}
Heckerman, D., Geiger, D., Chickering, D.M.: Learning {Bayesian} {Networks}: {The} {Combination} of {Knowledge} and {Statistical} {Data}. Machine Learning  \textbf{20}(3),  197--243 (Sep 1995)

\bibitem{howard1960}
Howard, R.A.: Dynamic Programming and Markov Processes. MIT Press, Cambridge, MA (1960)

\bibitem{Jensen_Nielsen}
Jensen, F.V., Nielsen, T.D.: Bayesian Networks and Decision Graphs. Springer New York, 2nd edn. (2007)

\bibitem{Kajita2020}
Kajita, S., Kinjo, T., Nishi, T.: Autonomous molecular design by Monte-Carlo tree search and rapid evaluations using molecular dynamics simulations. Communications Physics  \textbf{3}(1) (May 2020)

\bibitem{Kjaerulff_Madsen}
Kjaerulff, U.B., Madsen, A.L.: Bayesian Networks and Influence Diagrams: A Guide to Construction and Analysis. Springer Publishing Company, 2nd edn. (2013)

\bibitem{Kocsis2006}
Kocsis, L., Szepesvári, C.: Bandit Based Monte-Carlo Planning, p. 282–293. Springer Berlin Heidelberg (2006)

\bibitem{Koller_Friedman}
Koller, D., Friedman, N.: Probabilistic Graphical Models: Principles and Techniques - Adaptive Computation and Machine Learning. The MIT Press (2009)

\bibitem{Labbe2020}
Labbe, Y., Zagoruyko, S., Kalevatykh, I., Laptev, I., Carpentier, J., Aubry, M., Sivic, J.: Monte-Carlo Tree Search for Efficient Visually Guided Rearrangement Planning. IEEE Robotics and Automation Letters  \textbf{5}(2),  3715–3722 (Apr 2020)

\bibitem{leurent20a}
Leurent, E., Maillard, O.A.: Monte-Carlo Graph Search: the Value of Merging Similar States. In: Pan, S.J., Sugiyama, M. (eds.) Proceedings of The 12th Asian Conference on Machine Learning. Proceedings of Machine Learning Research, vol.~129, pp. 577--592. PMLR (18--20 Nov 2020)

\bibitem{li18aHC}
Li, A., van Beek, P.: Bayesian Network Structure Learning with Side Constraints. In: Kratochvíl, V., Studený, M. (eds.) Proceedings of the Ninth International Conference on Probabilistic Graphical Models. Proceedings of Machine Learning Research, vol.~72, pp. 225--236. PMLR (11--14 Sep 2018)

\bibitem{McLachlan2020}
McLachlan, S., Dube, K., Hitman, G.A., Fenton, N.E., Kyrimi, E.: Bayesian networks in healthcare: Distribution by medical condition. Artificial Intelligence in Medicine  \textbf{107},  101912 (Jul 2020)

\bibitem{Mo2022}
Mo, S., Pei, X., Wu, C.: Safe Reinforcement Learning for Autonomous Vehicle Using Monte Carlo Tree Search. IEEE Transactions on Intelligent Transportation Systems  \textbf{23}(7),  6766–6773 (Jul 2022)

\bibitem{Perez2015}
Perez, D., Mostaghim, S., Samothrakis, S., Lucas, S.M.: Multiobjective Monte Carlo Tree Search for Real-Time Games. IEEE Transactions on Computational Intelligence and AI in Games  \textbf{7}(4),  347–360 (Dec 2015)

\bibitem{FGES:2017}
Ramsey, J., Glymour, M., Sanchez-Romero, R., Glymour, C.: A million variables and more: the Fast Greedy Equivalence Search algorithm for learning high-dimensional graphical causal models, with an application to functional magnetic resonance images. International Journal of Data Science and Analytics  \textbf{3},  121 -- 129 (2017)

\bibitem{Scanagatta2019}
Scanagatta, M., Salmerón, A., Stella, F.: A survey on Bayesian network structure learning from data. Progress in Artificial Intelligence  \textbf{8}(4),  425–439 (May 2019)

\bibitem{bnlearn}
Scutari, M.: Learning Bayesian Networks with the bnlearn {R} Package. Journal of Statistical Software  \textbf{35}(3),  1–22 (2010)

\bibitem{Sevinc2020}
Sevinc, V., Kucuk, O., Goltas, M.: A Bayesian network model for prediction and analysis of possible forest fire causes. Forest Ecology and Management  \textbf{457},  117723 (Feb 2020)

\bibitem{Silver2016}
Silver, D., Huang, A., Maddison, C.J., et~al.: Mastering the game of Go with deep neural networks and tree search. Nature  \textbf{529}(7587),  484–489 (Jan 2016)

\bibitem{Silver2018}
Silver, D., Hubert, T., Schrittwieser, J., et~al.: A general reinforcement learning algorithm that masters chess, shogi, and Go through self-play. Science  \textbf{362}(6419),  1140–1144 (Dec 2018)

\bibitem{spirtes93}
Spirtes, P., Glymour, C., Scheimes, R.: Causation, Prediction and Search. Springer- Verlag, New York, USA (1993)

\bibitem{Weng2021}
Weng, D., Chen, R., Zhang, J., Bao, J., Zheng, Y., Wu, Y.: Pareto-Optimal Transit Route Planning With Multi-Objective Monte-Carlo Tree Search. IEEE Transactions on Intelligent Transportation Systems  \textbf{22}(2),  1185–1195 (Feb 2021)

\bibitem{Xie2022}
Xie, X., Xie, B., Xiong, D., Hou, M., Zuo, J., Wei, G., Chevallier, J.: New theoretical ISM-K2 Bayesian network model for evaluating vaccination effectiveness. Journal of Ambient Intelligence and Humanized Computing  \textbf{14}(9),  12789–12805 (Jul 2022)

\bibitem{wiechowski2022}
Świechowski, M., Godlewski, K., Sawicki, B., Mańdziuk, J.: Monte Carlo Tree Search: a review of recent modifications and applications. Artificial Intelligence Review  \textbf{56}(3),  2497–2562 (Jul 2022)

\end{thebibliography}

\end{document}